\documentclass[conference]{IEEEtran}
\IEEEoverridecommandlockouts                                  

\usepackage{todonotes}
\usepackage{graphicx}
\usepackage{capt-of}
\usepackage{float}
\usepackage{subcaption}
\usepackage{siunitx}
\usepackage{colortbl}
\usepackage{xcolor}
\definecolor{lightgray}{rgb}{0.83, 0.83, 0.83}
\usepackage{multirow}
\usepackage{booktabs}
\usepackage[normalem]{ulem}

\usepackage{amsmath,amssymb,amsfonts, mathtools,mathrsfs}
\newcommand{\colvec}[2][.9]{%
  \scalebox{#1}{%
    \renewcommand{\arraystretch}{1}%
    $\begin{bmatrix}#2\end{bmatrix}$%
  }
 }

\newtheorem{remark}{Remark}

\newcommand{\UAV}{$\alpha$-Ted6R }
\newcommand{\secstub}[1]{%
\begin{center}
  \noindent
  \newcount\i
  \i=0
  \loop
    \ifnum\i<#1
      .\\
      \advance\i by1
  \repeat
\end{center}
}

\usepackage{graphicx}
\usepackage{xcolor}

\title{\LARGE \bf
Towards safe and optimal flight: \\Viability Kernel MPC for Fully Actuated Multirotor\\
\thanks{*This work is partially supported by the European Union - NextGen-
erationEU: Mission 4 Component 2 - CUP C53D23000520006: project
STARLIT.}}

\author{\IEEEauthorblockN{Massimiliano Bertoni$^{1}$, Alberto Piccina$^{1}$, Gianni Lunardi$^2$, Elias Fontanari$^2$, \\ Andrea Del Prete$^2$, Angelo Cenedese$^{3,4}$ and Giulia Michieletto$^{1,3}$}
\IEEEauthorblockA{\textit{$^1$Department of Management and Engineering}, \textit{University of Padova},
Vicenza, Italy  \\
\textit{$^2$Industrial Engineering Department}, \textit{University of Trento},
Trento, Italy \\
\textit{$^3$Department of Information Engineering}, \textit{University of Padova},
Padova, Italy\\
\textit{$^4$Department of Industrial Engineering}, \textit{University of Padova},
Padova, Italy\\
corresponding contact: \texttt{massimiliano.bertoni@unipd.it}}
}

\begin{document}

\maketitle
\thispagestyle{empty}
\pagestyle{empty}

\begin{abstract}
Industrial aerial robotics demands safety guarantees for navigation in unstructured environments while optimizing performance and computational efficiency. This paper presents a method for generating safe pose trajectories for fully actuated multirotors within a Model Predictive Control (MPC) framework, leveraging both viability theory and data-driven methods. Obstacle avoidance is enforced through dynamically computed axis-aligned bounding boxes, providing formal safety guarantees without exhaustive offline reachability analysis. Numerical simulations on a fully actuated tilted hexarotor validate the approach, demonstrating successful navigation in cluttered environments with real-time computational performance.
\end{abstract}



\section{Introduction}

Industry interest in aerial robotics is rapidly growing, driven by the potential to perform tasks in diverse and challenging environments. However, the deployment of multirotors in real-world applications remains largely limited to passive monitoring and surveillance in wide open areas. This limitation stems primarily from the lack of strong safety and security guarantees for aerial robots, which prevents their adoption in more complex scenarios such as navigation in cluttered environments or operation in proximity to other robots and human operators~\cite{mohsan2023unmanned}.

The literature devoted to multirotor safety can be broadly categorized into two main trends. The first addresses the \textit{intrinsic safety} and robustness of the platforms themselves, proposing methods for the detection and diagnosis of hardware (actuators, main structure, and sensors) damage and degradation, and potentially performance losses~\cite{adaika2025fault}. From a control perspective, fault-tolerant regulation techniques have been extensively studied, enabling systems to maintain stable operation even under partial component failures~\cite{fourlas2021survey,saied2023review}. The second trend focuses on the \textit{extrinsic safety} of platforms in terms of collision avoidance, with significant research devoted to planning and control architectures that ensure navigation in static and dynamically cluttered environments~\cite{tang2021systematic,rezaee2024comprehensive}. Common approaches include reactive methods based on potential fields, sampling-based planners, and optimization-based strategies that incorporate obstacle constraints directly into the control formulation~\cite{quan2020survey}.

Despite these advances, most existing approaches either ensure short-term collision avoidance without guaranteeing long-term viability, or rely on computationally expensive offline reachability analysis that does not scale well to high-dimensional nonlinear systems or dynamically changing environments. Moreover, safety is often treated as a constraint to be satisfied reactively, rather than as an intrinsic property embedded within the planning and control architecture.

Overcoming these limitations, this work addresses the problem of pose trajectory generation for fully actuated multirotors navigating in cluttered environments by introducing a safety framework grounded in \emph{viability theory}. In detail, building upon the viability-based safe MPC framework proposed in~\cite{lunardi2024receding}, we exploit viability theory to formally characterize the set of states from which the system can be safely driven to an equilibrium while satisfying state and input constraints.
To this end, we adopt the \textit{Viability-Boundary Optimal Control} (VBOC) approach introduced in~\cite{la2023vboc} to compute a conservative numerical approximation of the viability kernel for multirotor platforms. The approximation is obtained offline and encoded via a neural network for real-time evaluation. It explicitly accounts for propeller saturation and obstacle-induced constraints, the latter enforced as membership in a dynamically computed obstacle-free box, in contrast to~\cite{lunardi2024receding}. The approximated viability kernel is embedded into the MPC problem as a hard terminal constraint, requiring the predicted final state to belong to the viable set. This guarantees that each feasible solution contains a state from which a safe equilibrium can be reached, thereby enabling a safe abort mechanism in the presence of persistent infeasibilities, as in~\cite{lunardi2024receding}.

The main contributions of this work are therefore: $(i)$ the formalization of the numerical viability kernel approximation for a fully actuated multirotor navigating in cluttered environments, and $(ii)$ its integration into a terminally constrained MPC framework that guarantees a certified safe fallback state in case of control infeasibility.

The remainder of this paper is organized as follows. Section~\ref{sec:system_model} presents the multirotor dynamic model. Section~\ref{sec:viability_kernel} describes the computation of the viability kernel via VBOC and its neural network encoding. Section~\ref{sec:mpc} details the MPC formulation with terminal viability constraints and the safe abort mechanism. Section~\ref{sec:results} presents simulation results validating the approach, and Section~\ref{sec:conclusions} concludes the paper with a discussion of future research directions.

\noindent\textit{Notation}:
The symbols $\mathbb{R}$ and $\mathbb{N}$ denote the sets of real and natural numbers, respectively, and $\mathbb{R}_{>0}$ the set of strictly positive real numbers. Lowercase italic and bold letters represent scalars and (column) vectors and uppercase bold letters denote matrices. The symbols $\mathbf{0}$ and $\mathbf{I}$ denote the zero vector/matrix and the identity matrix, with dimension clear from the context, while $\mathbf{1}$ denotes a vector of ones. The notation $\preceq$ indicates component-wise inequality.

\section{Multirotor Model} 
\label{sec:system_model}

To describe the dynamic model of a fully actuated multi-rotor, we introduce the global inertial frame $\mathscr{F}_W$ (\textit{world frame}) and the local platform-fixed frame $\mathscr{F}_B$ (\textit{body frame}). The multirotor \textit{state} is defined by the vector
$
    \mathbf{x} = \colvec{ \mathbf{p}^{\top} \; \boldsymbol{\eta}^{\top} \; \mathbf{v}^{\top} \; \boldsymbol{\omega}^{\top}}^\top \in \mathcal{X} \subset \mathbb{R}^{12}
$
where $\mathbf{p} \in \mathbb{R}^3$ identifies the position of its center of mass in the world frame,  
$\boldsymbol{\eta} \in \mathbb{R}^3$ collects the Euler angles $(\phi, \theta, \psi)$ describing the orientation of the platform (i.e., of $\mathscr{F}_B$) with respect to $\mathscr{F}_W$, and $\mathbf{v}, \boldsymbol{\omega} \in \mathbb{R}^3$ denote the linear and angular velocities in $\mathscr{F}_W$ and body frame $\mathscr{F}_B$, respectively. 
For compactness, the state can be partitioned as $\mathbf{x} = \colvec{ \boldsymbol{\xi}^{\top} \; \boldsymbol{\nu}^{\top} }^{\top}$, where $\boldsymbol{\xi} = \colvec{ \mathbf{p}^{\top} \; \boldsymbol{\eta}^{\top} }^{\top} \in \mathbb{R}^6$ is  the pose component
and $\boldsymbol{\nu} = \colvec{ \mathbf{v}^{\top} \; \boldsymbol{\omega}^{\top} }^{\top} \in \mathbb{R}^6$ is the twist component.

The multirotor  \textit{control input} $\mathbf{u} = \colvec{ \omega_1^2 \; \dots \; \omega_{n}^2}^{\top} \in \mathcal{U} \subset \mathbb{R}^{n}$ is the vector stacking the squared (limited) spinning rates of the $n \geq 4$ rotors. The platform dynamics can be expressed compactly as $\dot{\mathbf{x}} = f(\mathbf{x}, \mathbf{u})$, and explicitly as
\begin{equation}
\label{eq:system_dynamics}
\dot{\mathbf{x}} = \colvec{\dot{\mathbf{p}} \\ \dot{\boldsymbol{\eta}} \\ \dot{\mathbf{v}} \\ \dot{\boldsymbol{\omega}}} = \colvec{ \mathbf{v} \\
\mathbf{T}(\boldsymbol{\eta})\, \boldsymbol{\omega} \\
 g\mathbf{e}_3 + m^{-1} \mathbf{R}(\boldsymbol{\eta}) \mathbf{F}\, \mathbf{u} \\
\mathbf{J}^{-1}(-\boldsymbol{\omega} \times \mathbf{J}\boldsymbol{\omega}) + \mathbf{J}^{-1} \mathbf{M}\, \mathbf{u} }
\end{equation}
where the rotation matrix $\mathbf{R}(\boldsymbol{\eta}) \in SO(3)$ and the Euler kinematic matrix $\mathbf{T}(\boldsymbol{\eta}) \in \mathbb{R}^{3 \times 3}$ depend on the adopted rotation sequence, and $m \in \mathbb{R}$ and $\mathbf{J} \in \mathbb{R}^{3 \times 3}$ are respectively the mass and the inertia of the platform. In~\eqref{eq:system_dynamics}, the term $g \mathbf{e}_3$ models the gravitational force, while $\mathbf{F}, \mathbf{M} \in \mathbb{R}^{3 \times 6}$ are the control force and moment allocation matrices that map the control input to the wrench components at the multirotor center of mass~\cite{perin2024star}.

\begin{remark}
Consistently with the discrete-time implementation of embedded digital control systems, the discrete-time approximation $\mathbf{x}_{k+1} = f(\mathbf{x}_k, \mathbf{u}_k)$, $k \in \mathbb{N}$, of continuous-time dynamics~\eqref{eq:system_dynamics} is adopted hereafter for both the MPC formulation and the viability kernel computation. 
\end{remark}

\section{Viability Kernel Computation}
\label{sec:viability_kernel}

In general, the viability kernel of a dynamical system is defined as the largest subset of the state space from which safety can be ensured, meaning that the system can remain within a predefined set of safe states indefinitely under an appropriate control strategy.
Formally, for a generic discrete-time dynamic system, 
\begin{equation}
\mathbf{x}_{k+1}=f\left(\mathbf{x}_{k}, \mathbf{u}_{k}\right), \quad \mathbf{x} \in \mathcal{X} \subset \mathbb{R}^{n}, \quad \mathbf{u} \in \mathcal{U} \subset \mathbb{R}^{m},
\end{equation}
where $\mathcal{X}$ and $\mathcal{U}$ are compact sets, the viability kernel $\mathcal{V}_o$ corresponds to the subset of $\mathcal{X}$ from which it is possible to maintain the state within $\mathcal{X}$ indefinitely, namely
\begin{equation}
\label{eq:viability_kernel}
\mathcal{V}_o = \left\{\mathbf{x}_{0} \in \mathcal{X} \mid \exists \{\mathbf{u}_{k} \in \mathcal{U}\}_{k \geq 0} : \mathbf{x}_{k} \in \mathcal{X}, \; \forall k\geq 0 \right\}
\end{equation}

In this work, we derive a conservative approximation of the viability kernel for a fully actuated multirotor by adopting the VBOC approach proposed in~\cite{la2023vboc}. This method addresses the main challenge posed by the nonlinearity of the system~\eqref{eq:system_dynamics} through a reformulation based on equilibrium reachability.

Let $\mathcal{S}=\{\mathbf{x} \in \mathcal{X} \mid \exists \mathbf{u} \in \mathcal{U}: \mathbf{x}=f(\mathbf{x}, \mathbf{u})\}$ be the set of all equilibrium states of the system.  According to the VBOC approach, a state is viable if and only if there exists a control sequence that maintains the state within $\mathcal{X}$ indefinitely and drives it to an equilibrium in $\mathcal{S}$.  This yields a refined viability kernel $\mathcal{V} \subseteq \mathcal{V}_o$, defined as the infinite-horizon backward reachable set of $\mathcal{S}$. Formally,
\begin{equation}
\begin{aligned}
\label{eq:viability_kernel_VBOC}
\mathcal{V} = \left\{ \mathbf{x}_0 \in \mathcal{V}_o \mid \exists \{\mathbf{u}_{k} \in \mathcal{U}\}_{k \geq 0} : \lim_{k \to \infty} \mathbf{x}_k \in \mathcal{S}\right\}
\end{aligned}
\end{equation}
which represents the largest positively control-invariant set ensuring convergence to $\mathcal{S}$, as detailed in~\cite{la2023vboc}.

Through VBOC, it is possible to compute a conservative approximation $\hat{\mathcal{V}} \subseteq \mathcal{V}$ of the viability kernel~\eqref{eq:viability_kernel_VBOC}. While this approximation is not necessarily control-invariant in general, it can still be exploited to provide safety guarantees within the MPC framework~\cite{lunardi2024receding}. Practically, VBOC identifies $\hat{\mathcal{V}}$ by computing its boundary $\partial \hat{\mathcal{V}}$, which represents a conservative estimation of the boundary $\partial \mathcal{V}$ of the viability kernel~\eqref{eq:viability_kernel_VBOC}. This is achieved by solving the following optimal control problem (OCP) over a finite horizon $M \in \mathbb{N}$:
\begin{align}
\label{eq:ocp_vboc}
\underset{\left\{\mathbf{x}_{k}\right\}_{0}^{M},\left\{\mathbf{u}_{k}\right\}_{0}^{M-1}}\max & \quad \mathbf{a}^{\top} \mathbf{x}_{0} \\
\text { subject to } & \quad  \mathbf{x}_{k+1}=f\left(\mathbf{x}_{k}, \mathbf{u}_{k}\right) \quad k=0\ldots M-1 \notag \\
& \quad \mathbf{x}_{k} \in \mathcal{X}, \;\; \mathbf{u}_{k} \in \mathcal{U} \quad \;  k=0\ldots M-1 \notag \\
& \quad \mathbf{S} \mathbf{x}_{0}=\mathbf{s} \notag \\
& \quad \mathbf{x}_{M}=\mathbf{x}_{M-1} \notag 
\end{align}

\begin{figure}[t]
    \centering
    \includegraphics[width=0.75\columnwidth]{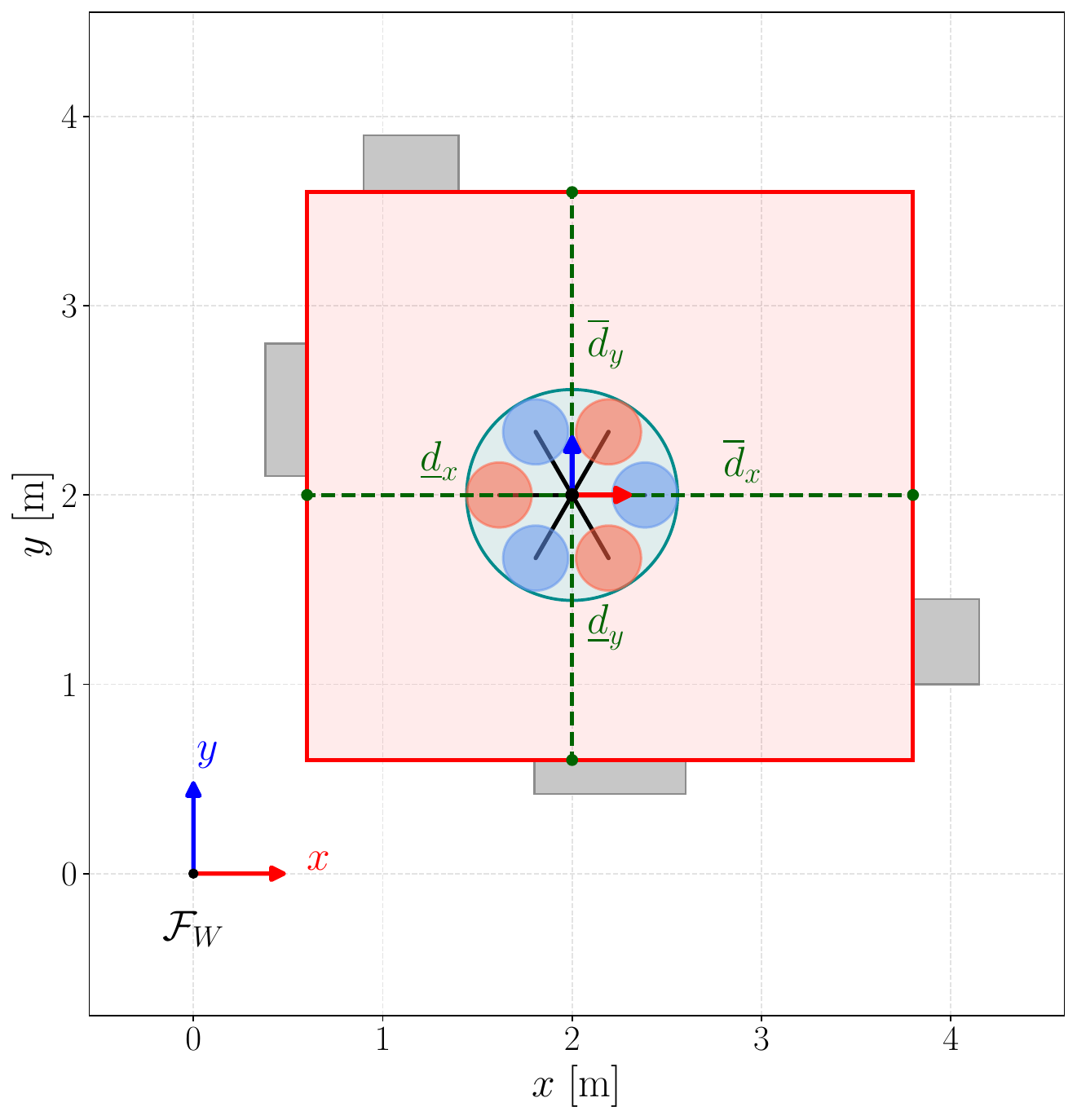}
    \caption{{Planar representative example of AABB (red area) in presence of obstacles (grey boxes).}}
    \label{fig:AABB}
\end{figure}

For a fully actuated multirotor, a convenient choice of  the cost vector $\mathbf{a} \in \mathbb{R}^{12}$, and of the initial constraint matrix and vector  $\mathbf{S} \in \mathbb{R}^{12\times 12}$, $\mathbf{s} \in \mathbb{R}^{12}$ is
\begin{align}
\mathbf{a}=\colvec{
\mathbf{0} \\
\mathbf{d}
}, \quad \mathbf{S}=\colvec{
\mathbf{I} & \mathbf{0} \\
\mathbf{0} & \mathbf{I}-\mathbf{d} \mathbf{d}^{\top}
}, \quad \mathbf{s}=\colvec{
\boldsymbol{\xi}_0 \\
\mathbf{0}}
\end{align}
where $\mathbf{d} \in \mathbb{R}^{6}$ is the unit vector identifying the twist direction, i.e., $\mathbf{d} = \boldsymbol{\nu} / \| \boldsymbol{\nu} \|_2$, and $\boldsymbol{\xi}_0 = \colvec{ \mathbf{p}^{\top}_0 \; \boldsymbol{\eta}^{\top}_0 }^{\top} \in \mathbb{R}^6$ is the initial pose. This choice of parameters yields the maximum initial twist norm along $\mathbf{d}$ as the solution of~\eqref{eq:ocp_vboc}. The resulting state trajectory also includes other states on $\partial \hat{\mathcal{V}}$, making the algorithm particularly efficient. 
As for the sets $\mathcal{X}$ and $\mathcal{U}$ in~\eqref{eq:ocp_vboc}, the input constraints are dictated by rotor spinning rate saturation, whereas the state constraints emerge from the presence of obstacles within the operating space. We assume that each element of the vector $\mathbf{u}$ is upper bounded by $\bar{u} \in \mathbb{R}_{> 0}$, i.e., $\mathcal{U} = \{\mathbf{u} \in \mathbb{R}^n \mid 0 \leq u_i \leq \bar{u}, \; i=1\ldots n\}$. We model the obstacle-free region surrounding the UAV via an Axis-Aligned Bounding Box (AABB).
Specifically, an AABB is a parallelepiped around the position $\mathbf{p}_0$ whose faces are parallel to the coordinate planes of $\mathscr{F}_W$, without loss of generality (Figure~\ref{fig:AABB}). According to the obstacle placement, an AABB is fully characterized by the distances between the origin of $\mathscr{F}_B$ and the AABB faces, namely by the vector $\mathbf{b} = \colvec{\underline{d}_{yz} \; \overline{d}_{yz} \; \underline{d}_{xz} \; \overline{d}_{xz} \; \underline{d}_{xy} \; \overline{d}_{xy}} \in \mathbb{R}^6$, where $\underline{d}_{ij}$ and $\overline{d}_{ij}$ denote the minimum and maximum distances from the face parallel to the ${ij}$-plane of $\mathscr{F}_W$, with $i,j \in \{x,y,z\}$. The constraint on the multirotor position can be formalized as $\mathbf{p} \in \mathcal{B}(\mathbf{p}_0, \mathbf{b}, r) \subset \mathbb{R}^3$, where

\begin{equation}
\mathcal{B}(\mathbf{p}_0, \mathbf{b}, r) = \left\{ \mathbf{p} \in \mathbb{R}^3 \;\middle|\;
\begin{aligned}
&(\mathbf{p}_0 + \underline{\mathbf{b}}) + r\,\mathbf{1} \preceq \mathbf{p} \\
&\mathbf{p} \preceq (\mathbf{p}_0 + \overline{\mathbf{b}}) - r\,\mathbf{1}
\end{aligned}
\right\}
\end{equation}
with $\underline{\mathbf{b}} = \colvec{\underline{d}_x \;\underline{d}_y\; \underline{d}_z}^{\top}$ and $\overline{\mathbf{b}} = \colvec{\overline{d}_x \; \overline{d}_y\; \overline{d}_z}^{\top}$, and $r \in \mathbb{R}_{> 0}$ is a safety margin representing the radius of a sphere within which the platform is fully inscribed. In other words, $\mathcal{B}(\mathbf{p}_0, \mathbf{b}, r)$ is the region obtained by eroding the original AABB by $r$ in all directions, ensuring collision-free motion.

In the resolution of~\eqref{eq:ocp_vboc}, we set the initial position $\mathbf{p}_0$ to zero, and we simulate different distances from the environment boundaries by varying $\underline{\mathbf{b}}$ and $\overline{\mathbf{b}}$ across OCP instances. 
In detail, following~\cite{la2023vboc}, we adopt a uniform random sampling of $\mathbf{b}$ (with $\underline{d}_{ij} \leq -r$ and $\overline{d}_{ij} \geq r$ for all $i,j \in \{x,y,z\}$ to exclude a priori infeasible cases), of initial pose $\boldsymbol{\xi}_0 = [\mathbf{p}_0^\top \;\boldsymbol{\eta}_0^{\top}]^{\top}$ and of twist direction $\mathbf{d}$ producing a dataset for training a Neural Network (NN) as a regression model. 
Adopting a data-driven approach, we encode $\mathcal{V}$ through a function $\phi(\boldsymbol{\eta}, \mathbf{d}, \mathbf{b}): \mathbb{R}^3 \times \mathbb{R}^{6} \times \mathbb{R}^{6} \rightarrow \mathbb{R}$ that maps Euler angles $\boldsymbol{\eta}$, twist direction $\mathbf{d}$ and AABB parameters $\mathbf{b}$ to the maximum viable twist norm $\gamma$. 
That is, if $\gamma=\phi(\boldsymbol{\eta}, \mathbf{d}, \mathbf{b})$, then $\colvec{\boldsymbol{\xi}^\top\; \gamma \mathbf{d}^\top}^\top \in \partial \mathcal{V}$. The adoption of a NN provides a convenient way to embed the viability check within each MPC iteration in a computationally efficient manner. 

\begin{remark}
The number of steps $M$ of the trajectory computed by the OCP~\eqref{eq:ocp_vboc} constitutes an upper bound on the number of time steps needed to safely drive the system to an equilibrium from $\hat{\mathcal{V}}$ when adopting a MPC strategy.
\end{remark}

\section{Terminal-Constraint MPC with Safe Abort}
\label{sec:mpc}

The ability to verify state viability is the key concept underlying the safe MPC framework, originally proposed in~\cite{lunardi2024receding}. We investigate this approach for a fully actuated multirotor. 

Given a target state $\mathbf{x}^r \in \mathbb{R}^{12}$ and an initial state condition $\mathbf{x}_0$, we  formulate the following MPC optimization problem
\begin{align}
\label{eq:mpc_formulation}
\underset{\left\{\mathbf{x}_{k}\right\}_{0}^{N},\left\{\mathbf{u}_{k}\right\}_{0}^{N-1}}{\min} & \quad \sum_{i=0}^{N-1} \ell_{k}\left(\mathbf{x}_{k}, \mathbf{u}_{k}\right)+\ell_{N}\left(\mathbf{x}_{N}\right)  \\
\text{subject to} & \quad \mathbf{x}_{k+1}=f\left(\mathbf{x}_{k}, \mathbf{u}_{k}\right) \quad k=0\ldots N-1  \notag \\
& \quad \mathbf{x}_{k} \in \mathcal{X}, \;\; \mathbf{u}_{k} \in \mathcal{U} \quad \; k=0 \ldots N-1  \notag \\
& \quad \mathbf{x}_{N} \in \mathcal{X}_{N}
\end{align}
where, without loss of generality, stage cost is $\ell_k(\mathbf{x}_k, \mathbf{u}_k) = \|\mathbf{x}_k - \mathbf{x}^{r}\|_\mathbf{Q}^2 + \|\mathbf{u}_k \|_\mathbf{R}^2$, penalizing tracking error and control effort, and terminal cost is $\ell_N(\mathbf{x}_N) = \|\mathbf{x}_N - \mathbf{x}^{r}\|_\mathbf{P}^2$.

\subsection{MPC Terminal Constraint}

At each MPC iteration, the state constraint set is defined as $\mathcal{X} = \{\mathbf{x}   \mid \mathbf{p} \in \mathcal{B}^\star\}$, where $\mathcal{B}^\star=\mathcal{B}(\mathbf{p}_0,\mathbf{b}^\star,r)$ is the collision-free AABB computed via a greedy shrinking procedure, and the terminal constraint set is defined as
\begin{equation}
\label{eq:Xn}
\mathcal{X}_N = \left\{\mathbf{x}   \;\middle|\; \phi\left(\boldsymbol{\eta}, \mathbf{d}, \mathbf{b}^\star\right) \geq \|\boldsymbol{\nu}\|_2\right\} \cap \hat{\mathcal{V}}
\end{equation}


In contrast to~\cite{lunardi2024receding}, the state constraint is enforced as membership in an obstacle-free box $\mathcal{B}^\star$, which is computed via a greedy shrinking algorithm. Let $\mathcal{B}_0$ denote the largest feasible AABB centered on the multirotor, defined by initial bounds $\mathbf{b}_0$. At each iteration $h$ of the greedy shrinking algorithm, let $\mathcal{F}_h=\{F_{h,1} \ldots F_{h,6}\}$ denote the set of the faces of $\mathcal{B}_h$ that intersect with obstacles. For each face $F_{h,p} \in \mathcal{F}_h$, let $d_{p}$ be the minimum displacement required to eliminate all collisions associated with that face. The face $F_{h,p^*} = \arg\max_{F_{h,p} \in \mathcal{F}_h} \text{vol}(\mathcal{B}_h^p)$ is selected, where $\mathcal{B}_h^{p}$ denotes the AABB obtained by displacing face $F_{h,p}$ by $d_{p}$ inward (i.e., shrinking the box). The updated box is then $\mathcal{B}_{h+1} = \mathcal{B}_h^{p^*}$. The procedure is repeated until $\mathcal{F}_h = \emptyset$, i.e., no faces intersect obstacles, yielding the collision-free AABB $\mathcal{B}^\star$ with parameters $\mathbf{b}^\star$. This greedy strategy prioritizes computational efficiency over global optimality, making it well-suited for real-time MPC replanning, though it may yield sub-optimal box configurations in geometrically complex environments.

\subsection{Safe Abort Procedure}

The hard constraint on the terminal state of the prediction horizon ($\mathbf{x}_{N} \in \mathcal{X}_{N}$ with $\mathcal{X}_{N}$ as in~\eqref{eq:Xn}) enforces membership in the viability kernel at each MPC iteration. If this constraint is satisfied, the predicted trajectory contains a state within the viability kernel, guaranteeing that, in the case of consecutive MPC infeasibilities, the system can reach a safe state and execute a predefined abort maneuver.  Specifically, upon detection of an infeasible MPC solution, the last feasible control sequence is applied, and a counter $c \in \mathbb{N}$ is initialized to zero and incremented at each subsequent time step. This counter tracks the number of steps since the last feasible solution and measures the steps remaining until the planned viable terminal state $\mathbf{x}_N^{\star}$ (from the last feasible solution) is reached. If a feasible solution is found before the counter reaches $N$ (i.e., $c < N$), the counter is reset to zero and MPC resumes normal operation. If the counter reaches the MPC horizon (i.e., $c = N$), the viable terminal state is reached, and the safe abort maneuver is triggered. Such a maneuver corresponds to the solution of the following OCP where $\mathbf{x}_{0} =  \mathbf{x}_N^{\star}$
\begin{align}
\underset{\left\{\mathbf{x}_{k}\right\}_{0}^{M},\left\{\mathbf{u}_{k}\right\}_{0}^{M-1}}{\min} & \sum_{i=0}^{M-1} \ell_{k}\left(\mathbf{x}_{k}, \mathbf{u}_{k}\right)+\ell_{M}\left(\mathbf{x}_{M}\right) \\
\text{subject to} \quad & \mathbf{x}_{k+1}=f\left(\mathbf{x}_{k}, \mathbf{u}_{k}\right) \quad  k=0\dots M-1 \notag\\
& \mathbf{x}_{k} \in \mathcal{X},\;\; \mathbf{u}_{k} \in \mathcal{U} \quad \; k=0\dots M-1 \notag\\
& \mathbf{x}_{M} = \mathbf{x}_{M-1} \notag
\end{align}



\section{Numerical Validation}
\label{sec:results}

\begin{figure}[t]
    \centering
    \includegraphics[width=\columnwidth]{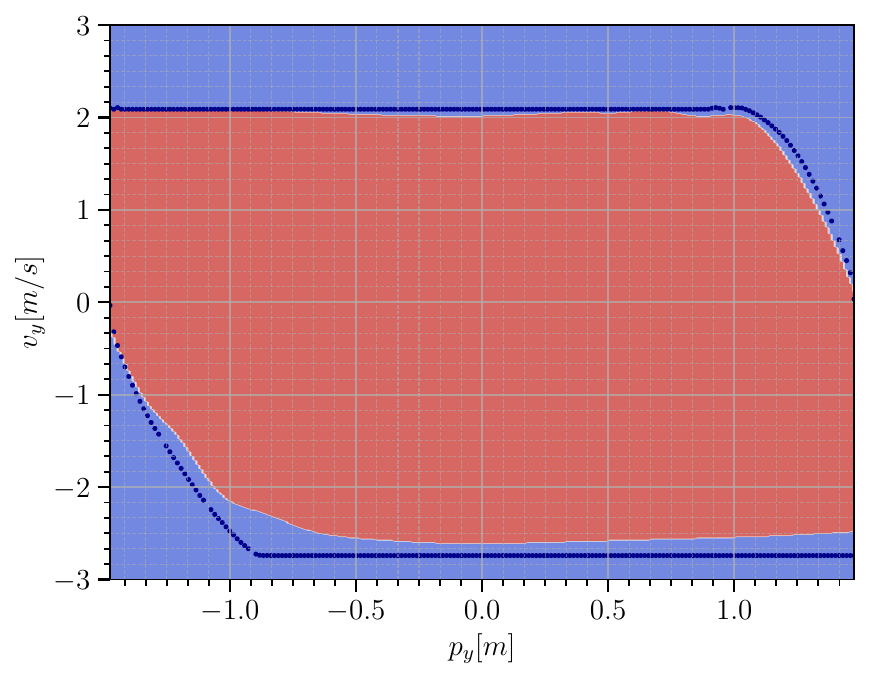}
    \caption{Viability kernel for the \UAV along only $y$-axis. Background color represents the outcomes of NN with red area identifying the viable set, black dots are the solution of the OCP~\eqref{eq:ocp_vboc}}
    \label{fig:vboc}
\end{figure}

\subsection{Simulation setup}

\paragraph{Application scenario}
The multirotor is tasked to reach a target pose $\boldsymbol{\xi}^{r} = \colvec{2 \; 0 \; 5 \; 0 \; 0\; 0}^{\top}$ with zero velocity, starting from the initial state $\mathbf{x}^0 = \mathbf{0}_{12}$, while avoiding the static obstacles whose center position $\mathbf{c}_o \in \mathbb{R}^3$ in $\mathscr{F}_W$ and whose dimensions $\ell_{o,x}, \ell_{o,y}, \ell_{o,z} \in \mathbb{R}_{>0}$ are defined in Table~\ref{tab:obstacles}.

\begin{table}[t]
    \centering
    \caption{Obstacle parameters.}
    \label{tab:obstacles}
    \begin{tabular}{ccccccc}
        \toprule
        & \multicolumn{3}{c}{center [m]} & \multicolumn{3}{c}{dimensions [m]} \\
        \cmidrule(lr){2-4} \cmidrule(lr){5-7}
        obstacle & $c_{o,x}$ & $c_{o,y}$ & $c_{o,z}$ & $\ell_{o,x}$ & $\ell_{o,y}$ & $\ell_{o,z}$ \\
        \midrule
        1 & $2.0$  & $0.0$ & $0.5$ & $0.5$ & $2.0$ & $2.0$ \\
        2 & $-1.5$ & $0.0$ & $1.0$ & $0.5$ & $3.0$ & $3.0$ \\
        3 & $0.0$  & $0.0$ & $3.0$ & $2.0$ & $2.0$ & $0.5$ \\
        \bottomrule
    \end{tabular}
\end{table}

\paragraph{Fully Actuated Multirotor} We carry out the validation process accounting for a coplanar star-shaped, interdependent cant-tilted hexarotor (hereafter referred to as $\alpha$-Ted6R) as the use case platform. According to~\cite{cigarini2026rotorsuitematlabsimulinktoolboxtilt}, this is a multirotor actuated by six propellers whose centers are evenly spaced on a circumference and whose spinning axes are tilted equally and in opposite directions along the arm axis for adjacent propellers, enabling independent control of all six degrees of freedom and thus achieving full actuation. The \UAV considered in this work is characterized by the parameters listed in Table~\ref{tab:uav_params}. The multirotor dynamics is computed with a time step of $0.1$ ms with a Runga-Kutta-4 integration.

\begin{table}[t]
    \centering
    \caption{UAV platform parameters.}
    \label{tab:uav_params}
    \begin{tabular}{cllc}
        \toprule
        $m$     & $[\text{kg}]$  & mass                       & $3.500$\\
        $\mathbf{J}$   & $[\text{kg\,m}^2]$                    & inertia matrix  & $\text{diag}(0.155,\, 0.147,\, 0.251)$       \\
        $l$        & $[\text{m}]$                                  & arm length                 & $0.385$                                   \\
                $r_{\text{p}}$    & $[\text{m}]$      & propeller radius           & $0.172$                                       \\
                       ${r}$    & $[\text{m}]$    & safe radius       & $0.557$    \\
        $\bar{u}$     & $[\text{Hz}^2]$          & max squared spinning rate    & $108^2$                       \\
        \bottomrule
    \end{tabular}
\end{table}

\paragraph{Control Framework}
The NN used to evaluate the viability constraint in the MPC formulation consists of two hidden layers, each with 512 units, using the GeLu activation function. The network was trained on a dataset of 200,000 samples. 
The MPC is implemented in Python using CasADi~\cite{andersson2019casadi} and ACADOS~\cite{verschueren2022acados}. The cost matrices are $\mathbf{Q} = \text{diag}(10^2, 10^2, 10^2,\, 10^2, 10^2, 10^2,\, 10^1, 10^1, 10^1,\, 5\cdot10^1, 5\cdot10^1, 5\cdot10^1)$, $\mathbf{R} = \mathbf{I}$, $\mathbf{P} =  \mathbf{Q}$. We impose the horizon $N = 60$ and the sampling period $T_s = 20$ ms.
The overall duration of simulation is fixed to $5$ s.

\begin{figure*}[t!]
    \centering
    \begin{subfigure}[b]{0.32\linewidth}
        \centering
        \includegraphics[trim={3.5cm 1.0cm 2.0cm 1.5cm},clip,width=\linewidth]{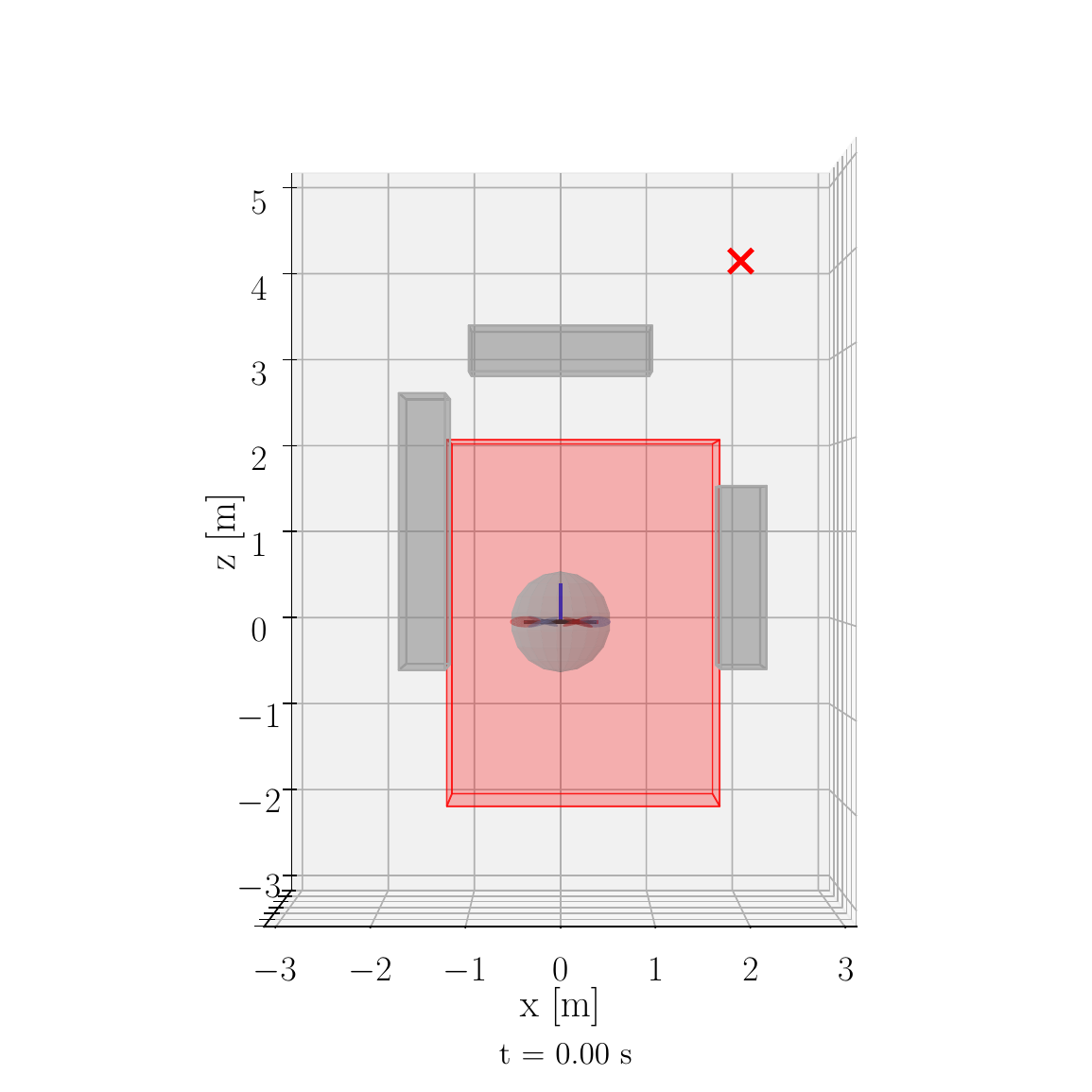}
        \caption{$t=0.00$s}
    \end{subfigure}%
    \begin{subfigure}[b]{0.32\linewidth}
        \centering
        \includegraphics[trim={3.5cm 1.0cm 2.0cm 1.5cm},clip,width=\linewidth]{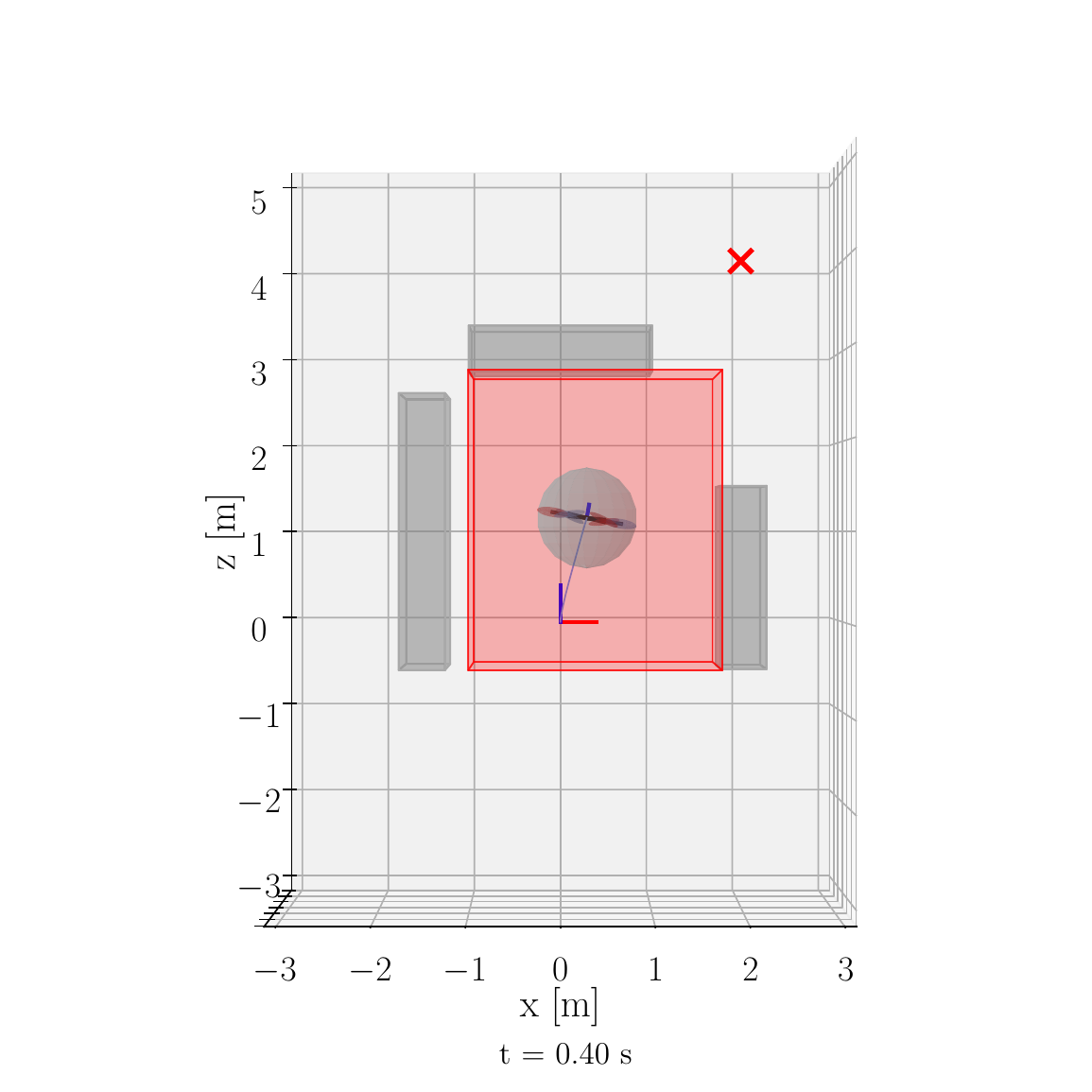}
        \caption{$t=0.40$s}
    \end{subfigure}%
    \begin{subfigure}[b]{0.32\linewidth}
        \centering
        \includegraphics[trim={3.5cm 1.0cm 2.0cm 1.5cm},clip,width=\linewidth]{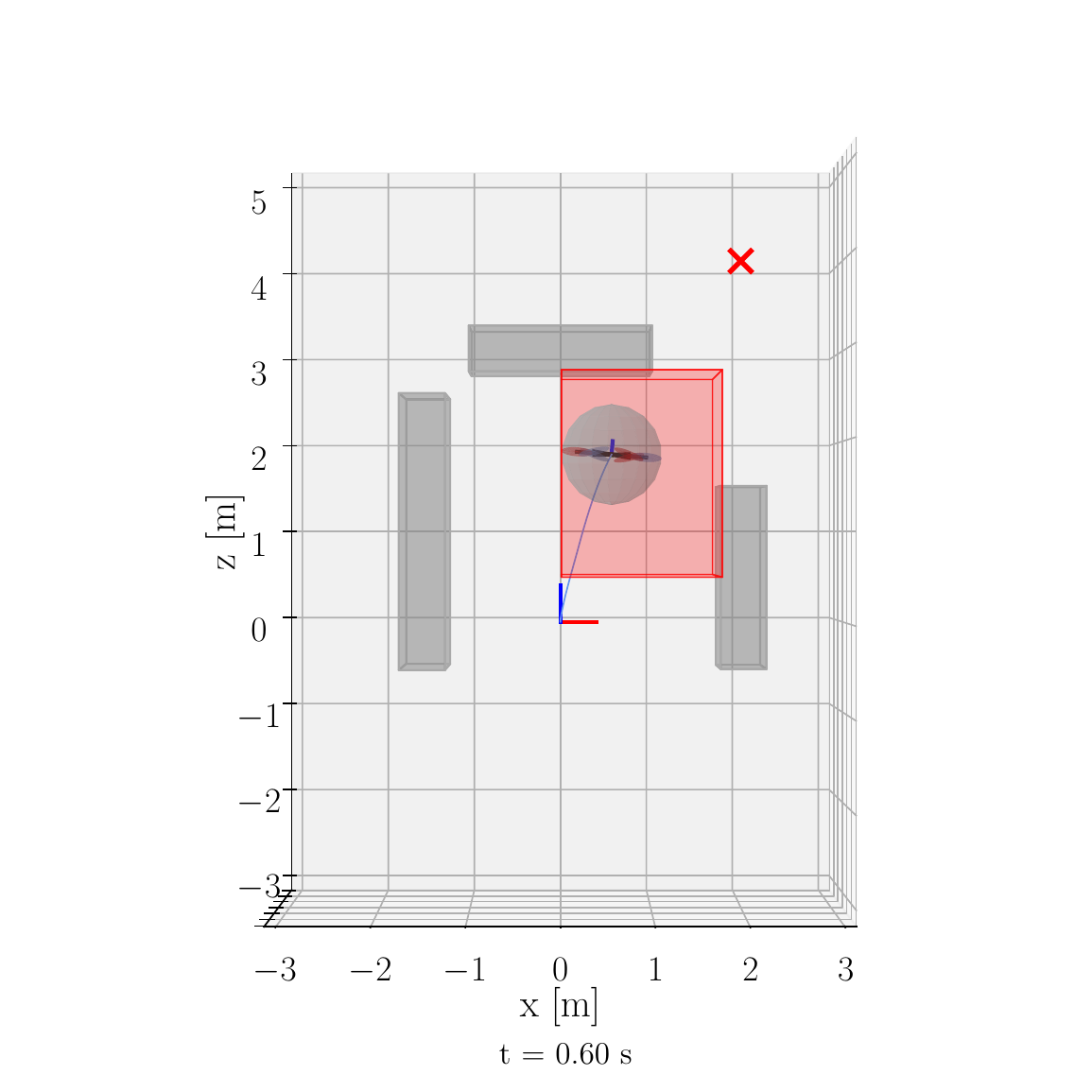}
        \caption{$t=0.60$s}
    \end{subfigure}
    
    \vspace{0.2cm} 
    
    \begin{subfigure}[b]{0.32\linewidth}
        \centering
        \includegraphics[trim={3.5cm 1.0cm 2.0cm 1.5cm},clip,width=\linewidth]{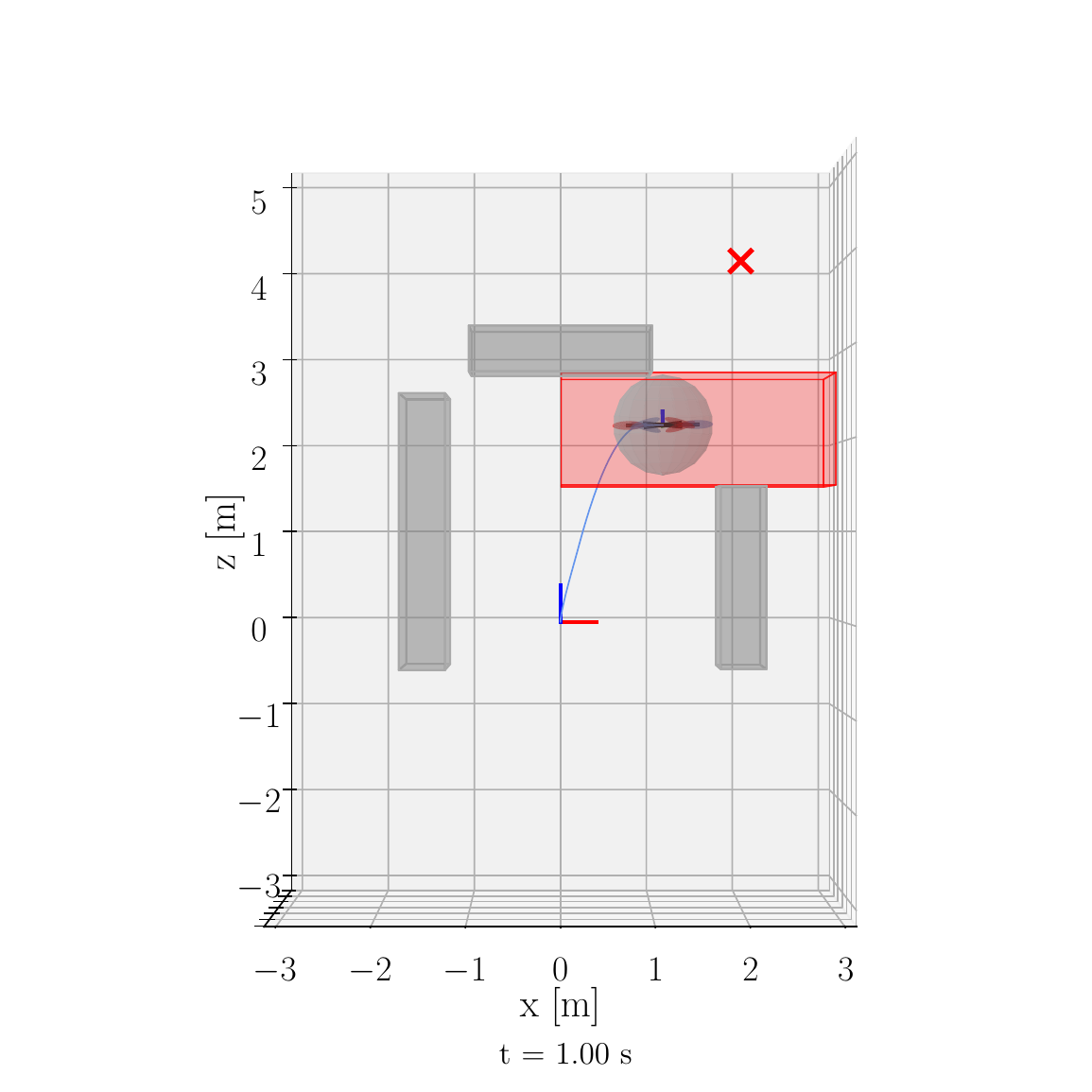}
        \caption{$t=1.00$s}
    \end{subfigure}%
    \begin{subfigure}[b]{0.32\linewidth}
        \centering
        \includegraphics[trim={3.5cm 1.0cm 2.0cm 1.5cm},clip,width=\linewidth]{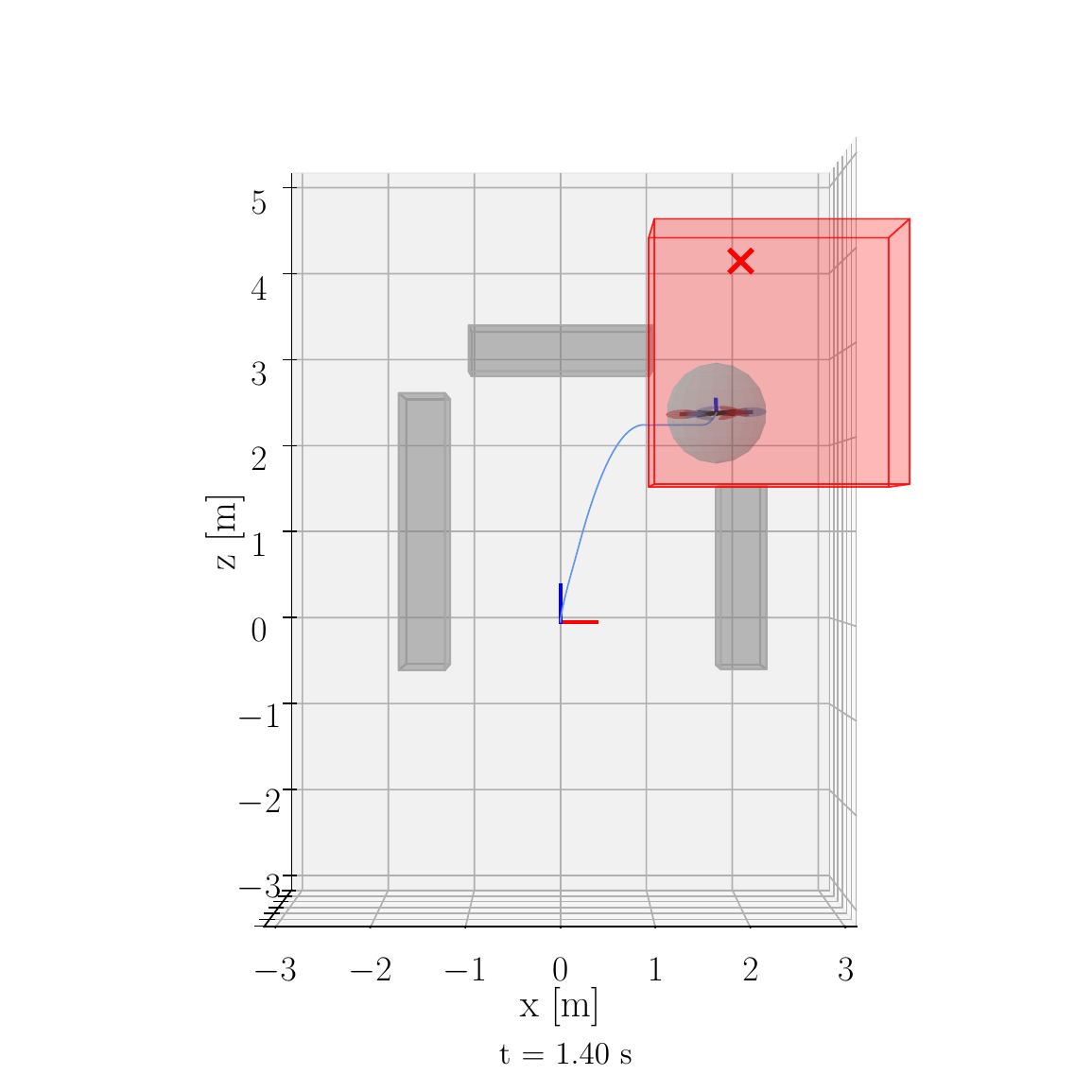}
        \caption{$t=1.40$s}
    \end{subfigure}%
    \begin{subfigure}[b]{0.32\linewidth}
        \centering
        \includegraphics[trim={3.5cm 1.0cm 2.0cm 1.5cm},clip,width=\linewidth]{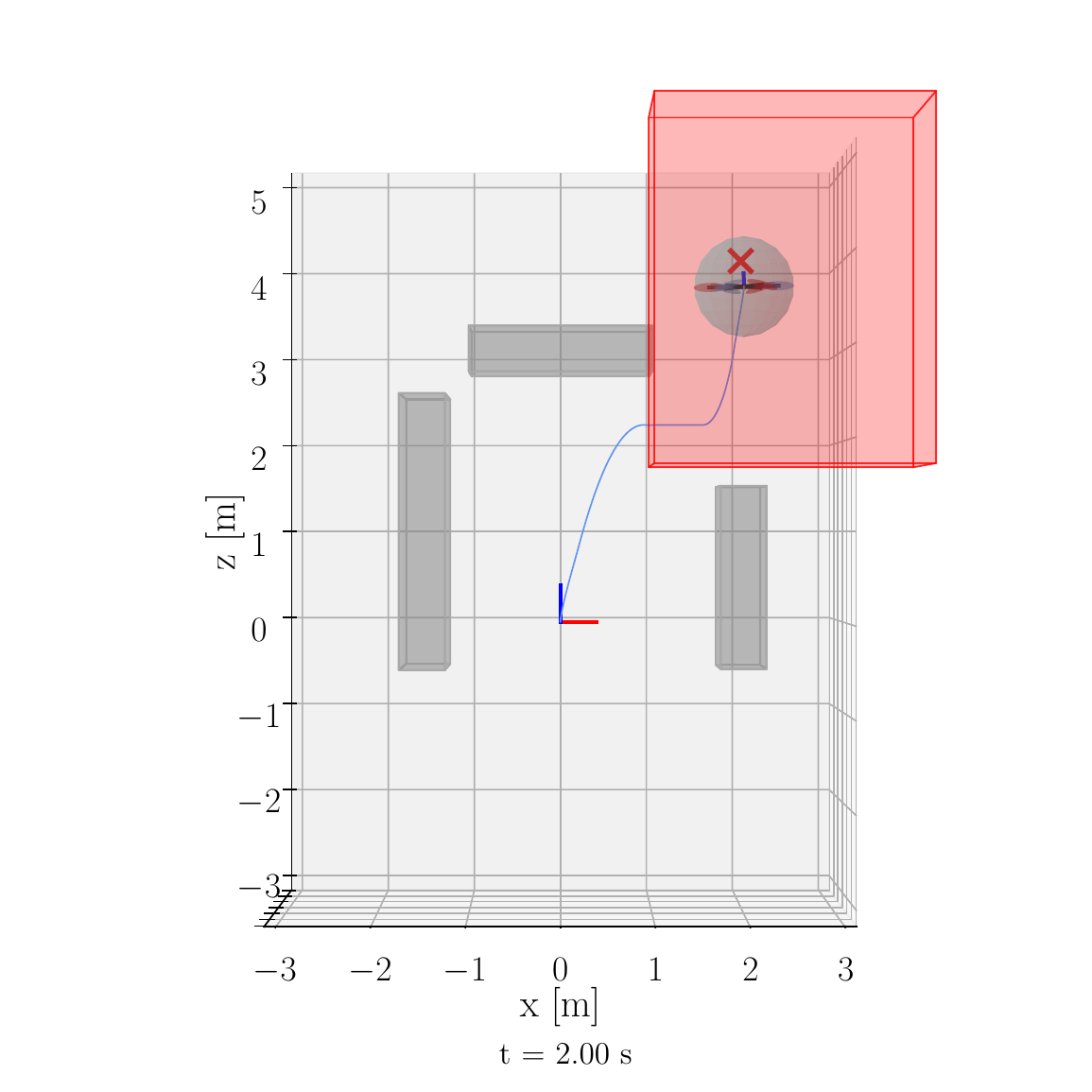}
        \caption{$t=2.00$s}
    \end{subfigure}

 \caption{Pose trajectory tracking performance of the \UAV using the terminal-constraint MPC with safe abort. 3Dview: red boxes indicate the time-varying AABBs, while gray blocks represent obstacles.}
    \label{fig:snapshots}
\end{figure*}

\subsection{Results}

\paragraph{NN Architecture} 
To illustrate the NN effectiveness, Figure~\ref{fig:vboc} compares the solution of~\eqref{eq:ocp_vboc} (black dots)  with the NN prediction (shown in red) for motion restricted to the $y$-axis. In this scenario, the twist direction reduces to $\mathbf{d} = [0, \pm1, 0, 0, 0, 0]^{\top}$, the initial orientation is $\boldsymbol{\eta}_0 = \mathbf{0}$, and the AABB bounds are $\mathbf{b}=\colvec{  0 \; 0 \;  \underline{d}_y \;  \overline{d}_y  \;  0 \; 0}^{\top}$ with $\underline{d}_y, \overline{d}_y \in [-2, 2]$ m uniformly sampled. The OCP optimization horizon is set to $M=15$.
Remarkably, the region predicted by the NN closely matches the OCP solutions, validating the NN's ability to accurately encode the viability kernel approximation $\hat{\mathcal{V}}$. 

\begin{figure*}[t]
    \centering
    \begin{subfigure}[b]{0.48\linewidth}
        \centering
        \includegraphics[width=\linewidth]{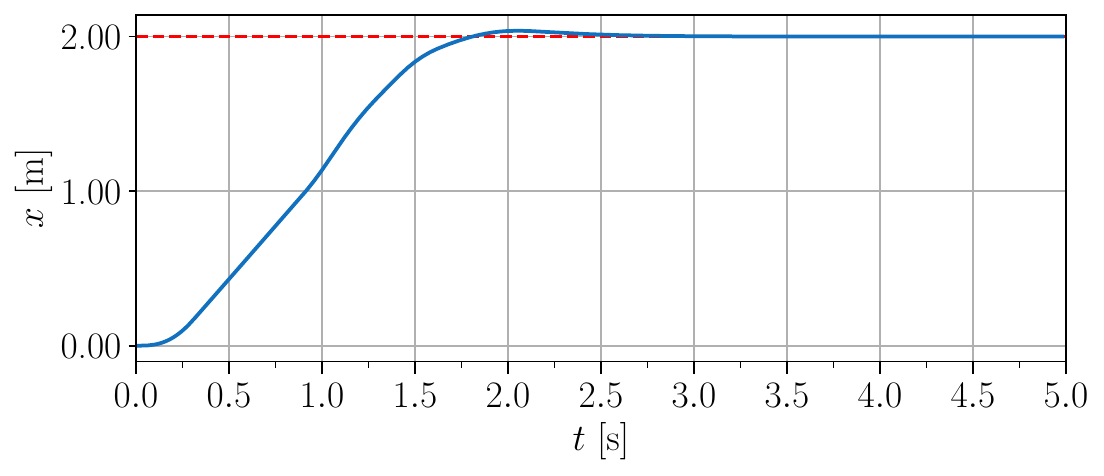} \\
        \includegraphics[width=\linewidth]{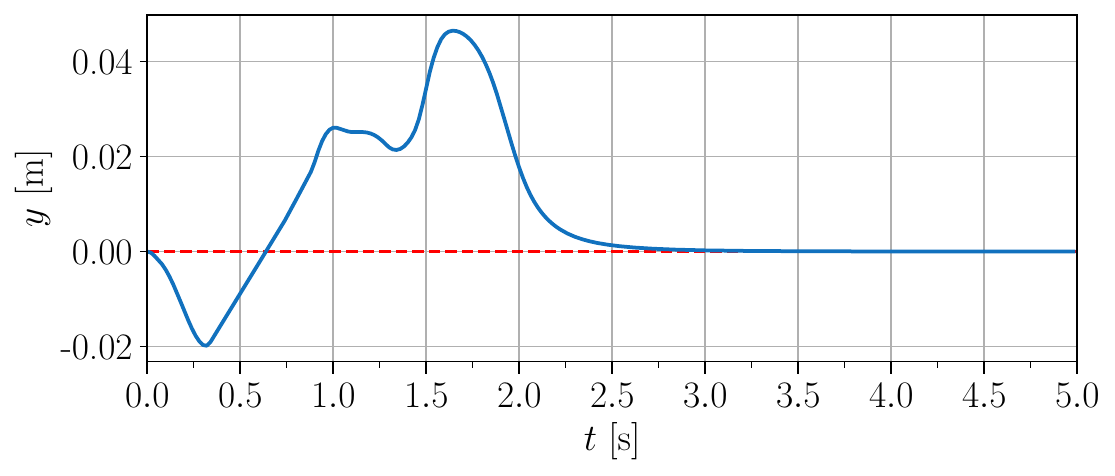} \\
        \includegraphics[width=\linewidth]{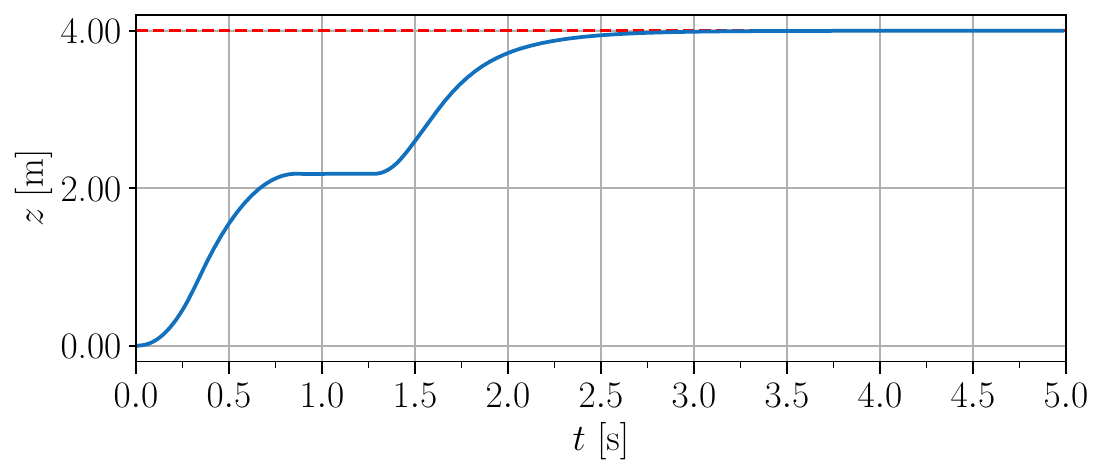}
        \caption{Position}
    \end{subfigure}
    \hfill
    \begin{subfigure}[b]{0.48\linewidth}
        \centering
        \includegraphics[width=\linewidth]{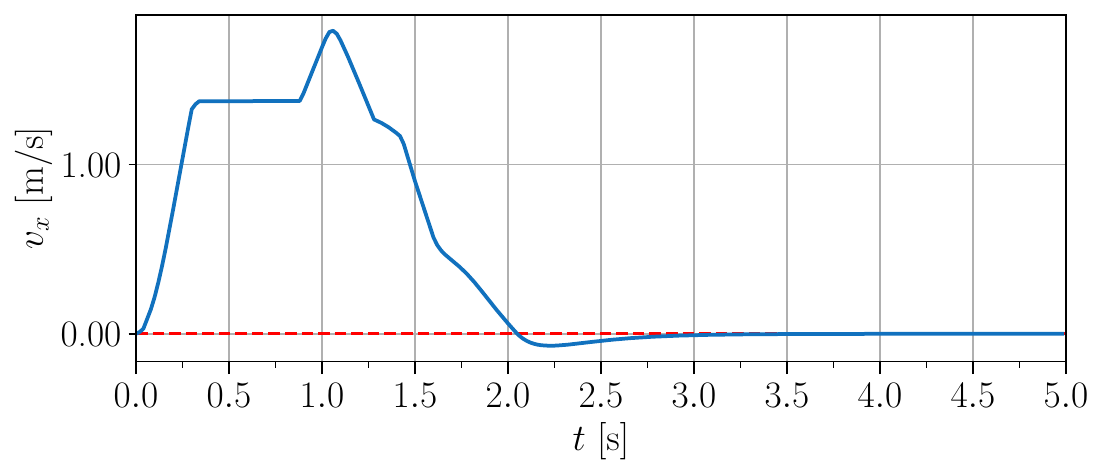} \\
        \includegraphics[width=\linewidth]{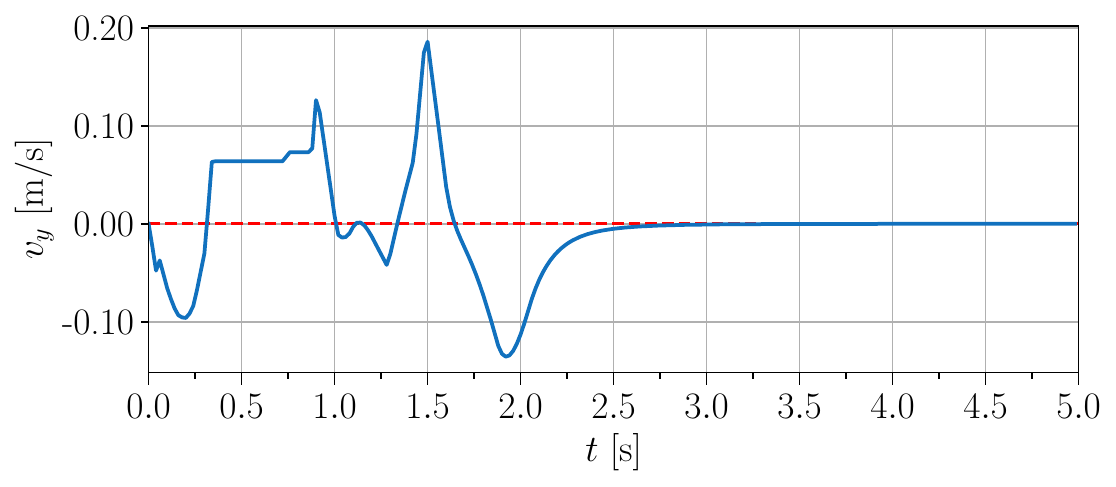} \\
        \includegraphics[width=\linewidth]{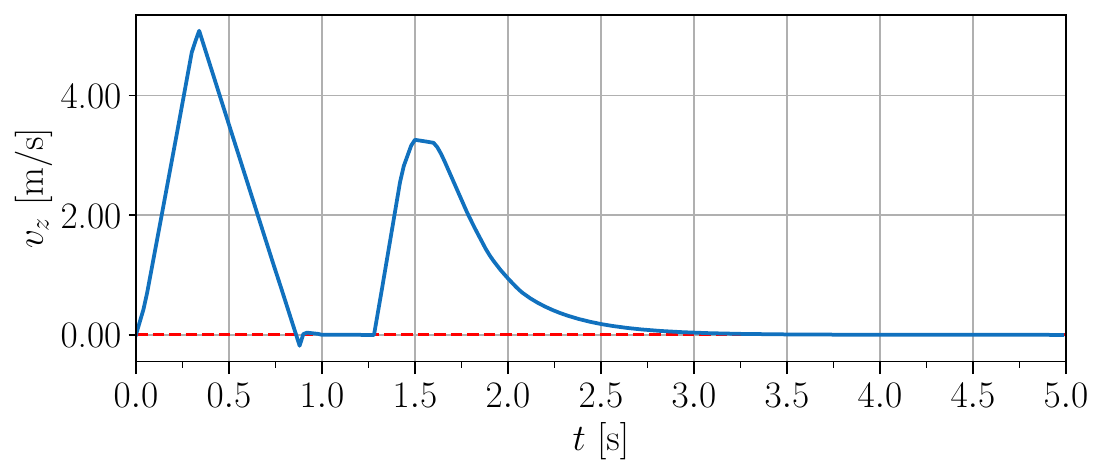}
        \caption{Linear velocity}
    \end{subfigure}

    \caption{Pose trajectory tracking performance of the \UAV using the terminal-constraint MPC with safe abort. States' evolution (blue) over the reference (red).}
    \label{fig:trends}
\end{figure*}

\paragraph{MPC Framework}
As can be seen from Figure~\ref{fig:snapshots}, the target pose is successfully reached despite not initially being contained within the obstacle-free box (limited to $\pm2$ m per axis). As \UAV navigates, new regions are progressively discovered and the AABB is dynamically updated via the greedy shrinking algorithm. Figure~\ref{fig:trends} reports the \UAV position and linear velocity evolution. 
The temporary plateau in the $z$ component during $[0.8, 1.4]$ s occurs when the safety sphere (radius $r$) becomes tangent to an AABB face, limiting vertical motion. During this phase, the MPC compensates by increasing the velocity along the $x$-axis.

Computational performance is excellent: average iteration time is $1.00$ ms (max $11.72$ ms, min $0.12$ ms), consistently below the imposed sampling period. This efficiency stems from the neural network encoding of the viability kernel and the tractable greedy AABB algorithm, enabling real-time operation.

\section{Conclusions}
\label{sec:conclusions}

This work presented a methodological proof of concept for safe pose trajectory tracking of fully actuated multirotors in cluttered environments. The proposed architecture combines viability theory with receding-horizon control to provide safety guarantees under state and input constraints, without relying on exhaustive offline reachability analysis.

As a foundational study, this work serves as a starting point that paves the way for multiple future investigations in different directions, including the extensions to more complex environments populated by dynamical obstacles and the integration of advanced perception or learning-based components within the safety framework.
From a technical point of view, future work aims to explore more accurate representations of obstacle-free regions, reducing the conservativeness of the current box-based approach and investigating computational strategies that integrate optimization with safety considerations that would otherwise require exhaustive searches, thereby enabling more efficient and flexible real-time planning.

\bibliographystyle{IEEEtran}
\bibliography{references.bib}

@article{la2023vboc,
  title={Vboc: Learning the viability boundary of a robot manipulator using optimal control},
  author={La Rocca, Asia and Saveriano, Matteo and Del Prete, Andrea},
  journal={IEEE Robotics and Automation Letters},
  volume={8},
  number={11},
  pages={6971--6978},
  year={2023},
  publisher={IEEE}
}

@inproceedings{lunardi2024receding,
  title={Receding-Constraint Model Predictive Control using a Learned Approximate Control-Invariant Set},
  author={Lunardi, Gianni and La Rocca, Asia and Saveriano, Matteo and Del Prete, Andrea},
  booktitle={2024 IEEE International Conference on Robotics and Automation (ICRA)},
  pages={11626--11632},
  year={2024},
  organization={IEEE}
}

@inproceedings{perin2024star,
  title={Star-shaped Tilted Hexarotor Maneuverability: Analysis of the Role of the Tilt Cant Angles},
  author={Perin, Marco and Bertoni, Massimiliano and Viezzer, Nicolas and Michieletto, Giulia and Cenedese, Angelo},
  booktitle={2024 IEEE 20th International Conference on Automation Science and Engineering (CASE)},
  pages={1791--1797},
  year={2024},
  organization={IEEE}
}

@article{andersson2019casadi,
  title={CasADi: a software framework for nonlinear optimization and optimal control},
  author={Andersson, Joel AE and Gillis, Joris and Horn, Greg and Rawlings, James B and Diehl, Moritz},
  journal={Mathematical Programming Computation},
  volume={11},
  number={1},
  pages={1--36},
  year={2019},
  publisher={Springer}
}

@article{verschueren2022acados,
  title={acados—a modular open-source framework for fast embedded optimal control},
  author={Verschueren, Robin and Frison, Gianluca and Kouzoupis, Dimitris and Frey, Jonathan and Duijkeren, Niels van and Zanelli, Andrea and Novoselnik, Branimir and Albin, Thivaharan and Quirynen, Rien and Diehl, Moritz},
  journal={Mathematical Programming Computation},
  volume={14},
  number={1},
  pages={147--183},
  year={2022},
  publisher={Springer}
}

@article{mohsan2023unmanned,
  title={Unmanned aerial vehicles ({UAVs}): Practical aspects, applications, open challenges, security issues, and future trends},
  author={Mohsan, Syed Agha Hassnain and Othman, Nawaf Qasem Hamood and Li, Yanlong and Alsharif, Mohammed H and Khan, Muhammad Asghar},
  journal={Intelligent service robotics},
  volume={16},
  number={1},
  pages={109--137},
  year={2023},
  publisher={Springer}
}

@article{adaika2025fault,
  title={Fault detection and diagnosis methodologies for unmanned aerial vehicles: State-of-the-art},
  author={Adaika, Zineb and Al-Haddad, Luttfi A and Giernacki, Wojciech and Jaber, Alaa Abdulhady and Boumehraz, Mohamed and Hamzah, Mohsin N and Flayyih, Mujtaba A},
  journal={Journal of Intelligent \& Robotic Systems},
  volume={111},
  number={2},
  pages={63},
  year={2025},
  publisher={Springer}
}

@article{fourlas2021survey,
  title={A survey on fault diagnosis and fault-tolerant control methods for unmanned aerial vehicles},
  author={Fourlas, George K and Karras, George C},
  journal={Machines},
  volume={9},
  number={9},
  pages={197},
  year={2021},
  publisher={MDPI}
}

@article{saied2023review,
  title={A review on recent development of multirotor UAV fault-tolerant control systems},
  author={Saied, Majd and Shraim, Hassan and Francis, Clovis},
  journal={IEEE Aerospace and Electronic Systems Magazine},
  volume={39},
  number={9},
  pages={146--180},
  year={2023},
  publisher={IEEE}
}

@article{tang2021systematic,
  title={Systematic review of collision-avoidance approaches for unmanned aerial vehicles},
  author={Tang, Jun and Lao, Songyang and Wan, Yu},
  journal={IEEE Systems Journal},
  volume={16},
  number={3},
  pages={4356--4367},
  year={2021},
  publisher={IEEE}
}

@article{rezaee2024comprehensive,
  title={Comprehensive review of drones collision avoidance schemes: Challenges and open issues},
  author={Rezaee, Mohammad Reza and Hamid, Nor Asilah Wati Abdul and Hussin, Masnida and Zukarnain, Zuriati Ahmad},
  journal={IEEE Transactions on Intelligent Transportation Systems},
  volume={25},
  number={7},
  pages={6397--6426},
  year={2024},
  publisher={IEEE}
}

@misc{cigarini2026rotorsuitematlabsimulinktoolboxtilt,
      title={RotorSuite: A MATLAB/Simulink Toolbox for Tilt Multi-Rotor UAV Modeling}, 
      author={Nicola Cigarini and Giulia Michieletto and Angelo Cenedese},
      year={2026},
      eprint={2602.18814},
      archivePrefix={arXiv},
      primaryClass={cs.RO},
      url={https://arxiv.org/abs/2602.18814}
}

@article{quan2020survey,
  title={Survey of UAV motion planning},
  author={Quan, Lun and Han, Luxin and Zhou, Boyu and Shen, Shaojie and Gao, Fei},
  journal={IET Cyber-systems and Robotics},
  volume={2},
  number={1},
  pages={14--21},
  year={2020},
  publisher={Wiley Online Library}
}

\end{document}